\documentclass[a4paper,UKenglish,cleveref, autoref, thm-restate]{tgdk-v2021}

\pdfoutput=1 

\footnotetext{Equal contribution}

\footnotetext{Corresponding author}

\title{Standardizing Knowledge Engineering Practices with a Reference Architecture}

\titlerunning{Reference Architectures for Knowledge Engineering}

\author{Bradley P. Allen$^\dagger$}{University of Amsterdam, Amsterdam, The Netherlands \and \url{https://www.bradleypallen.org} }{b.p.allen@uva.nl}{https://orcid.org/0000-0003-0216-3930}{}

\author{Filip Ilievski$^{\dagger*}$}{Vrije Universiteit, Amsterdam, The Netherlands \and \url{http://www.ilievski.info} }{ilievski@isi.edu}{https://orcid.org/0000-0002-1735-0686}{}

\authorrunning{B.\,P. Allen and F. Ilievski}

\Copyright{Bradley P. Allen and Filip Ilievski}

\category{Position}



\relatedversion{} 

\nolinenumbers 

\SeriesVolume{2}
\SeriesIssue{1}
\SectionAreaEditor{Aidan Hogan, Ian Horrocks, Andreas Hotho, and Lalana Kagal}
\SpecialIssue{Trends in Graph Data and Knowledge -- Part 2}
\DateSubmission{}
\DateAcceptance{}
\DatePublished{}
\ArticleNo{5}

\begin{document}

\maketitle

\begin{abstract}
Knowledge engineering is the process of creating and maintaining knowledge-producing systems. Throughout the history of computer science and AI, knowledge engineering workflows have been widely used given the importance of high-quality knowledge for reliable intelligent agents. Meanwhile, the scope of knowledge engineering, as apparent from its target tasks and use cases, has been shifting, together with its paradigms such as expert systems, semantic web, and language modeling. The intended use cases and supported user requirements between these paradigms have not been analyzed globally, as new paradigms often satisfy prior pain points while possibly introducing new ones. The recent abstraction of systemic patterns into a boxology provides an opening for aligning the requirements and use cases of knowledge engineering with the systems, components, and software that can satisfy them best, however, this direction has not been explored to date. This paper proposes a vision of harmonizing the best practices in the field of knowledge engineering by leveraging the software engineering methodology of creating reference architectures. We describe how a reference architecture can be iteratively designed and implemented to associate user needs with recurring systemic patterns, building on top of existing knowledge engineering workflows and boxologies. We provide a six-step roadmap that can enable the development of such an architecture, consisting of scope definition, selection of information sources, architectural analysis, synthesis of an architecture based on the information source analysis, evaluation through instantiation, and, ultimately, instantiation into a concrete software architecture. We provide an initial design and outcome of the definition of architectural scope, selection of information sources, and analysis. As the remaining steps of design, evaluation, and instantiation of the architecture are largely use-case specific, we provide a detailed description of their procedures and point to relevant examples. We expect that following through on this vision will lead to well-grounded reference architectures for knowledge engineering, will advance the ongoing initiatives of organizing the neurosymbolic knowledge engineering space, and will build new links to the software architectures and data science communities. 
 \end{abstract}

\section{Introduction}



Knowledge engineering (KE) is the collection of activities for eliciting, capturing, conceptualizing, and formalizing knowledge to be used in information systems~\cite{sabouknowledge}. KE includes two broader tasks of creating and maintaining knowledge~\cite{allen2023identifying}.
Throughout the history of computer science and AI, KE workflows have been a critical component when building reliable intelligent agents across domains and tasks~\cite{li2023trustworthy}. Indeed, it has been intuitive that developing trustworthy models for applications, from common sense to traffic, crime, and weather, requires well-understood knowledge processes \cite{lenat2023getting}. Similarly, task solutions within these domains, including question answering, summarization, and forecasting, are expected to incorporate standardized KE procedures to be meaningfully applicable and compatible with humans~\cite{tiddi2023knowledge}.



The importance of knowledge production processes has yielded many notable architectures over the past decades, which aim to synthesize dominant patterns and best practices. As apparent from these architectures,
the dominant paradigm of KE has been shifting through the history of AI, from its early days through the eras of expert systems and semantic web.
Expert system workflows, like CommonKADS~\cite{schreiber2000knowledge}, enable the extraction of expert knowledge into knowledge bases based on lifecycle analysis and corresponding models. Many Semantic Web applications can be aligned to a general layered template, most famously the Semantic Web Layer Cake and its contemporary variants~\cite{goebel2008role,hendler2009tonight,beek2016lod,lan2022semantic}. 
Knowledge graph (KG) workflows~\cite{tamavsauskaite2022defining} and comprehensive toolkits~\cite{ilievski2020kgtk} aim to bridge the gap between knowledge bases and their applications, showing a larger emphasis on extensional and possibly less precise modeling of knowledge~\cite{simsek2022knowledge}.
Knowledge graph engineering (KGE) emerged as a variant of KE geared towards capturing, representing, and utilizing complex information about entities, their relationships, and their underlying semantics~\cite{simsek2022knowledge,groth2023knowledge}.
Researchers and domain experts have devised KGE workflows that are tailored to the needs of a variety of domains like biomedicine~\cite{lobentanzer2022democratising}, library and information sciences~\cite{tharani2021much}, web democracy~\cite{tommasini}, commonsense knowledge~\cite{ilievski2021cskg}, and publications~\cite{priem2022openalex}. Analogously, enterprise infrastructures, such as the Amazon Product Knowledge Graph~\cite{zalmout2021all}, have been devised for commercial settings, without clear reference to a standardized workflow. 
The recent trends in KE, such as the consideration of large language models (LLMs) as knowledge artifacts~\cite{petroni2019language} and the prominence of neurosymbolic systems~\cite{van2019boxology,sabouknowledge}, bring a new perspective to KE. The role of LLMs in KE workflows is studied actively to understand the potential of LLMs to enhance, replace, or add KE components~\cite{guan2023leveraging}. 
Meanwhile, recent work based on abstracting semantic web and machine learning systems (SWeMLS)~\cite{ekaputra2023describing} indicated the prominence of KE in neurosymbolic systems, with around a quarter of all SWeMLS patterns corresponding to a KE process. 

The present landscape of KE methodologies and tools lacks a comprehensive framework of user needs and available paradigms, as each subsequent KE era does not necessarily include the benefits brought by its predecessors~\cite{allen2023identifying}.
KE systems require a principled way of considering different user requirements, paradigms, and use cases, thus combining human, social, and technical factors~\cite{hogan2020semantic}. 
To address this need, recent work has proposed the formalism of a \textit{boxology}: a hierarchical taxonomy of systemic design patterns expressed in a graphical notation (cf. \autoref{fig:RA-synthesis-from-patterns})~\cite{van2019boxology,sabouknowledge}. Boxologies provide an opening for aligning the requirements and use cases of knowledge engineering with the systems, components, and software that can satisfy them best, however, this direction has not been explored to date. We see an urgency to understand the scope and purpose of KE in the latest evolving AI landscape, aiming to devise a general framework characterized by requirement-driven best practices. Such a framework should ideally support the perspectives of existing and emerging KE paradigms, enable a flexible definition and support for stakeholder requirements and priorities, and build on top of prior work on KE workflows and systemic patterns. Given the dynamic nature of KE, it should also provide mechanisms to adapt to evolving requirements over time and across applications. It should provide a prescriptive framework that standardizes best practices while allowing them to be customized to specific circumstances.

This paper proposes a vision of harmonizing the field of knowledge engineering by leveraging the software engineering methodology of devising a reference architecture (RA), inspired by successful RAs designed and applied in domains such as the automotive sector, e-government, and service-oriented solutions~\cite{garces2021three}. RAs serve as a framework that standardizes a community of practice through a software engineering artifact, based on a survey of relevant stakeholders, their requirements, existing community-based workflows, and suitable evaluation. RAs are developed in a human-centric and iterative manner, which makes them particularly suitable for dynamic and frequently changing disciplines such as KE. They provide a common framework, while simultaneously enabling users to design specific RAs for their narrow use cases. The proposal to develop a methodology for devising RAs for KE is in line with the suggestions in 
\cite{hogan2020semantic}: we hypothesize that the development of RAs will make KE and related knowledge technologies accessible for outsiders and newcomers from disciplines such as software engineering and data science that have compatible goals.

We consider how RAs can be iteratively designed and implemented to associate user needs with recurring systemic patterns, building on top of existing KE workflows and boxologies. 
Section~\ref{sec:refarch} provides an extended problem statement that motivates the need to consolidate KE practices by building on top of existing boxology patterns. Section~\ref{sec:rw} provides relevant background on existing RAs and common methodologies for their principled development, and reviews state-of-the-art architectures for KE. Section~\ref{sec:roadmap} details a six-step methodology to design and implement a human-centric reference architecture that standardizes KE practices, consisting of scope definition, selection of information sources, architectural analysis, synthesis of an architecture based on the information source analysis, evaluation through instantiation, and, ultimately, instantiation into a concrete software architecture. We describe an initial design of the architectural scope, information sources, and analysis, and prescribe the processes for the design, evaluation, and instantiation of the architecture, as the latter steps are highly use-case dependent. 
The paper is concluded in Section~\ref{sec:conc}.
We expect that following through on this vision will lead to well-grounded reference architectures for knowledge engineering, and will facilitate further links to the software architectures and data science communities.

\section{Why a Reference Architecture Framework for KE?}
\label{sec:refarch}

The pursuit of a general reference architecture framework for KE is motivated in this section by three key factors. First, the KE paradigms have been shifting over time, each following one addressing pain points of the existing paradigms, but often failing to address other requirements. Second, KE users vary greatly, and while the user needs have been often discussed in the context of a specific application, a broad view of connecting users, their tasks, and their corresponding requirements is lacking. Third, the emerging boxology of neurosymbolic systems, with its recent link to knowledge engineering, provides a unique opportunity to exploit emerging patterns as components of a more comprehensive architecture.


\subsection{Historiographic perspective: Consolidation of the KE paradigms}
\label{ssec:history}

Prior research on KE is rich, spanning from the 1950s, through the expert systems era of the 1980s, the Semantic Web era, and the recent view of language modeling as a knowledge production process~\cite{ramsey1929knowledge,newell1958elements,feigenbaum1977art,feigenbaum1992personal,schreiber2000knowledge,berners2001semantic,petroni2019language,hogan2020semantic,ilievski2020kgtk,bender2021dangers,hartig2022reflections,wqdssearchteam2022wqds,alkhamissi2022review}. These different periods have approached KE following the contemporary technological, scientific, and societal focus. We provide a brief historiographic view on KE here, for an extended discussion we refer the reader to \cite{allen2023identifying}.


In the 1960s, researchers like Newell and Simon~\cite{newell1958elements} were hopeful about the ability of goal-directed search with heuristics to perform practical general-purpose problem-solving. However, by the 1970s, it became evident that these systems were not easy to scale to complex applications. During the mid-1970s, Feigenbaum~\cite{feigenbaum1977art}, influenced by Newell and Simon's work, maintained that focusing on specific domains was crucial for successful knowledge engineering. Knowledge engineers worked on the elicitation of domain-specific knowledge with high quality and expressivity, and domain-independence and scalability were often not prioritized. This period saw a surge in creating expert systems for decision support in businesses, but by the early 1990s, it was clear that these systems had limitations, being brittle and hard to maintain. Efforts to address these limitations included the development of structured methodologies for knowledge engineering during the late 1990s~\cite{schreiber2000knowledge}. Feigenbaum~\cite{feigenbaum1992personal} persisted in exploring the idea of domain-specific applications but suggested that future systems should be scalable, globally distributed, and interoperable. These ideas, ahead of their time, foreshadowed some aspects of what later became the World Wide Web.  

In the era of the Semantic Web, Tim Berners-Lee~\cite{berners2001semantic} advocated the use of specific open standards (e.g., RDF and SPARQL) to encode knowledge in Web content, to improve access and discoverability of Web content, and to enable automated reasoning. However, adoption of Semantic Web technology was slow, ultimately leading researchers to seek ways to align these standards and principles more closely with general software industry norms and make them more developer-friendly. Recent efforts, particularly in commercial knowledge graphs developed by companies like Google and Amazon~\cite{zalmout2021all}, have shown a shift towards custom architectures, often based on property graphs. This shift, while innovative, often sidesteps the interoperability and federation ideals of early visionaries like Feigenbaum and Berners-Lee. As a result, there's a growing need to refine what KE offers developers, focusing on comprehensive, scalable, customizable, and modular infrastructures that integrate with common data formats. KE should be domain-independent, supporting a wide range of use cases with user-friendly interfaces.

The rise of connectionist methods and graphical processing hardware in the 2010s has introduced new possibilities for knowledge production using large language models (LLMs). LLMs have been shown to be a means to provide robustness to missing schema and better generalization across domains and knowledge types. Two main perspectives have emerged regarding the relationship between LLMs and knowledge bases~\cite{alkhamissi2022review}. The first sees LLMs as standalone, queryable knowledge bases that can learn from unstructured text with minimal human intervention~\cite{petroni2019language}. This method challenges traditional, labor-intensive KE processes, but raises concerns about accuracy, ethical use, interoperability, and curatability. The second, more cautious perspective views language models as components in a KE workflow, combining new and traditional methods~\cite{ilievski2020kgtk}. This approach emphasizes accessibility, manual editing of extracted knowledge, and explanation of reasoning methods, addressing the limitations of earlier technologies. Both perspectives highlight the importance of sustainability and affordability in KE processes.

\begin{table*}[t]
    \caption{Representative user tasks and scenarios for knowledge engineering.}
    \label{tab:ke_scenarios}
    \small
    \centering
    \renewcommand{\arraystretch}{1.5}
    \raggedright
    \begin{tabular}{p{4.2cm}|p{10cm}} 
        \textbf{Task} & \textbf{Scenario} \\ 
        \hline
        \textbf{Ontology creation} & A new domain is identified, for which an ontology needs to be created. \\ 
         \textbf{Ontology refinement} & A new concept or relationship is identified in the domain, and the ontology needs to be modified to support it without disruptions. \\ 
         \textbf{Data ingest and transformation} & Multiple data sources provide overlapping or complementary information. The system needs to transform and normalize this data to ensure consistency in the knowledge graph. \\
         \textbf{Data source integration} & A new data source, in a previously unsupported schema, needs to be incorporated into the knowledge graph while ensuring data quality. \\ 
         \textbf{Anomaly detection} & The system flags a potential inconsistency or contradiction in the knowledge graph, which needs to be resolved. \\ 
         \textbf{Knowledge graph completion} & The system flags a missing or incomplete statement, which needs to be automatically inserted. \\ 
         \textbf{Human oversight of knowledge graph quality} & A subject matter expert (SME) identifies a piece of outdated or incorrect information in the knowledge graph, which needs to be flagged to initiate a correction. \\ 
         \textbf{Human feedback} & As SMEs interact with the system, they might have insights or suggestions based on their domain expertise, which needs to be supported and incorporated into the refinement process. \\ 
    \end{tabular}

\end{table*}

\subsection{Social perspective: Systematic procedures for incorporating stakeholder tasks and needs} 
KE tasks can be roughly split into two main categories in terms of their goal: creating and maintaining knowledge artifacts. Here, the knowledge artifacts are typically knowledge graphs, ontologies, and taxonomies~\cite{sabouknowledge}. Representative KE tasks are shown in \autoref{tab:ke_scenarios}. These include tasks of creating an ontology or refining that ontology, for example, to include a newly discovered concept or relationship~\cite{noy2001ontology,fernandez1997methontology,suarez2011neon}. KE tasks include ingesting and transforming data from multiple data sources into a single artifact, or performing data integration between different schemas~\cite{jain2010ontology,dong2015schema}. Resulting KGs can be further refined to address issues such as inconsistencies of modeling, contradictions of factual information, outdated information, or missing/incomplete statements, to improve their quality~\cite{paulheim2017knowledge}. Such issues may be raised by automated systems~\cite{shenoy2021study} as well as human subject matter experts (SMEs)~\cite{piscopo2018models}. Finally, humans interacting with the knowledge artifact may have further suggestions or feedback for refinement~\cite{ilievski2020kgtk}.

Commonly, KE procedures identify local requirements for an artifact, with an implicit assumption of the user profile. Meanwhile, user studies are increasingly present in KE research~\cite{abian2022analysis,iglesias2023comparison}, a trend that can be enriched by considering the natural plurality of users. 
While early KE might have been carried out by computer science practitioners, today it often includes domain experts interacting with knowledge directly, knowledge engineers building ontologies, knowledge editors fixing outdated information, data scientists developing knowledge completion systems, and business/organizational stakeholders that stress-test the available knowledge to understand its value~\cite{iglesias2023comparison}.
Considering the example tasks in \autoref{tab:ke_scenarios}, we note that the tasks of ontology creation and data integration require expertise from knowledge engineers and subject matter experts. Refining of knowledge artifacts can be performed by knowledge engineers or data scientists, whereas providing human feedback and oversight requires subject matter experts. Many of these tasks may also benefit from the inputs from business and organizational stakeholders. While here we refer to \textit{stakeholders} as humans that create and maintain knowledge engineering processes, there is an additional set of tasks and users that \textit{use} the artifact resulting from the knowledge engineering process, such as data scientists developing AI prototypes that reason over knowledge and software developers that build knowledge-infused chatbots.


\subsection{Component perspective: Preliminary source of architectural patterns}

Driven by societal and technical needs for explainability, robustness, and collaboration, neurosymbolic AI has been growing in popularity recently, emerging as one of the key trends of AI research~\cite{kautz2022third,sabouknowledge}. Each neurosymbolic system combines neural, machine-learning components with symbolic manipulation. Given the breadth of this definition, there have been attempts to organize neurosymbolic systems by abstracting their architectures using recurring patterns. Following the \textit{boxology} approach~\cite{van2019boxology,sabouknowledge}, data structures can be symbols or data, whereas algorithmic modules are either inductive (machine learning) or deductive (based on knowledge representation and reasoning formalisms). Then, each boxology pattern is a combination of alternating data structure and algorithmic module boxes. The initial boxology work identified 15 such patterns. 11 of these, together with 33 new patterns, were found in the systems systematically surveyed in~\cite{breit2023combining}. The 44 patterns have been classified into a pattern typology based on their complexity, e.g., simple patterns have a single processing unit. Sample patterns from this boxology are illustrated in \autoref{fig:simple-swemls-patterns}, which we describe in more detail in \autoref{sec:roadmap}.

The question emerges: what is the role of KE in NeSy AI systems? How often do NeSy architectures represent KE processes? An insight into these questions is provided by \cite{sabouknowledge}, who coined the term \textit{neurosymbolic knowledge engineering} and analyzed the NeSy approaches that combine machine learning and semantic web components. While we see such component analysis as a step in the right direction of organizing neurosymbolic KE, we identify three challenges for state-of-the-art KE that are apparent from the boxology framework.
\textit{Challenge 1:} The patterns should be associated with user requirements, tasks, and application needs to enable their efficient and precise application. The present boxology patterns do not include this information, nor have the mechanism built in to include it in the future. \textit{Challenge 2:} Mechanisms for aligning with ongoing trends and shifting requirements are lacking. These are needed as the set of boxology patterns and their specification (e.g., type constraints) are still largely in flux, as apparent from the large number of newly discovered patterns in~\cite{breit2023combining}, and given the lack of specification of how the boxology primitives align with popular NeSy processes (e.g., fine-tuning) and artifacts (e.g., knowledge graphs). \textit{Challenge 3:} There is a lack of a standard for communicating the KE boxology patterns to non-knowledge engineers, including software engineers, data scientists, domain experts, and business stakeholders. While the abstraction of the boxology patterns makes a step towards facilitating broader comprehension, the patterns are still meant for experts.



\section{Related Work}
\label{sec:rw}

The previous section discussed three considerations that motivate the need for a reference architecture framework for the standardization of KE practices. 
This section summarizes definitions, practices, and methodologies associated with work on RAs, and describes how research in the areas of data, knowledge, and ontology engineering can contribute to the establishment of reference architectures for KE, towards the end of consolidating the three motivating perspectives. 


\subsection{Reference architectures}

\textbf{Definition and uses} A reference architecture is a framework that aligns stakeholders' requirements with design patterns through a final architecture and a corresponding software system. As such, an RA serves as a generic architecture for a class of information systems within a software engineering community of practice \cite{angelov2009classification}. 
RAs have several shared characteristics: they provide the highest level of abstraction, they emphasize heavily architectural qualities, their stakeholders are considered but absent from the architecture, they promote adherence to common standards, and are effective for system development and communication~\cite{ataei2022state}. Notably, while architectures capture software structures, not every structure is architectural: architecture is an abstraction that should emphasize the attributes that are important to stakeholders~\cite{bass2022software}. For a comprehensive review of software RAs, we refer the reader to~\cite{garces2021three}.

RAs are driven by two emerging trends~\cite{cloutier2010concept}. First, an increasing complexity, scope, and size of the system of interest, its context, and the organizations creating the system. Second, increasing dynamics and integration, i.e., shorter time to market, more interoperability, rapid changes, and adaptations in the field. In \cite{bass2022software}, the authors identify thirteen uses for developing a central reference architecture. A key aspect of reference architectures, along with related types of architectural artifacts, is that they are key to the creation of a technology strategy that drives consensus across multiple groups of stakeholders with an enterprise engaged in software application development for business purposes. Other benefits include that RAs enable the system's quality attributes, enable early prediction of system qualities, encode fundamental design decisions, support the training of new team members, reduce system complexity, and facilitate reuse. Reference architectures provide a common lexicon and taxonomy, a common architectural vision, and modularization~\cite{cloutier2010concept}. Notably, good architecture is necessary but not sufficient to ensure quality.

Many RAs have been proposed in the past decades, some of which have gained wide adoption in their domains. Well-known examples are AUTOSAR for automotive sector~\cite{staron2021autosar}, CORBA for object integration through brokers~\cite{boldt95}, S3 for service-oriented solutions~\cite{arsanjani2007s3}, EIRA for e-Government systems,\footnote{\url{https://joinup.ec.europa.eu/collection/european-interoperability-reference-architecture-eira/about}} and NIST's Big Data Interoperability Framework~\cite{framework2015draft}. We describe AUTOSAR in greater detail to provide an example of a typical RA. First introduced in 2003, AUTOSAR was developed as a cooperative effort between major automotive manufacturers, suppliers, and tool developers. The primary goal of AUTOSAR is to enable the development of highly modular, scalable, and reusable software components for automotive applications. By providing a common software infrastructure and standardized interfaces, AUTOSAR aims to reduce development costs, improve software quality, and facilitate the integration of software components from multiple suppliers. 


Ironically, while the field of Semantic Web puts a lot of emphasis on developing artifacts like ontologies and knowledge graphs that enable common understanding between humans and machines, it has not caught up on the idea of developing architectures, such as AUTOSAR, that will provide a common framework in which different concerns can be expressed, negotiated, and resolved among stakeholders for large, complex knowledge systems~\cite{bass2022software}.


\textbf{Methodologies for creating RAs} A method to design a software architecture has been proposed by~\cite{nakagawa2011aspect}, consisting of five steps: establishing its scope, selecting and investigating information sources, performing an architectural analysis to identify architecturally significant requirements, carrying out synthesis of the reference architecture, and evaluating the architecture through surveys as well as its instantiation and use. Typical RAs for big data usually follow a three-step lifecycle consisting of data ingestion, transformation, and serving~\cite{ataei2022state}. Their major architectural components can be roughly grouped into 1) big data management and storage, 2) data processing and application interfaces, and 3) big data infrastructure. Two types of requirements are commonly used to describe stakeholder needs for such software architectures: functional requirements (FRs) and quality attributes (QAs)~\cite{bass2022software}. \textit{Functional requirements} typically describe what the system components are responsible for, i.e., they state what the system must do and how it must behave or react to runtime stimuli. They are satisfied by assigning an appropriate sequence of responsibilities throughout the architectural design. \textit{Quality attribute} is ``a measure or testable property of a system that is used to indicate how well the system satisfies the needs of its stakeholders.'' Quality attributes must be characterized using one or more scenarios, and they must be unambiguous and testable. 

An example of the application of RAs to knowledge engineering is the work of Ocaño et al. on an RA for integrating artificial intelligence and knowledge bases to support journalists and newsrooms \cite{ocana2023software}. They apply a methodology similar to that of \cite{nakagawa2011aspect}, taking domain-specific requirements for the effective support of journalistic activities, defining a reference architecture, and then implementing a prototype instantiation of that architecture. This architecture provides a crucial example of what a realization of an RA for KE would look like for a particular domain. This paper provides a streamlined procedure for instantiating other procedures based on a general RA framework.
In the case of general RAs for KE, prior work has devised a set of QAs and FRs~\cite{allen2023identifying} based on a historiographic analysis. The present paper considers how this effort can be advanced to result in a general-purpose RA framework for KE.


\subsection{Methodologies and workflows for knowledge engineering}
Reference architectures can serve as a framework that shapes and optimizes knowledge engineering workflows, ensuring they are efficient, scalable, and compliant with best practices and standards; conversely, knowledge engineering methodologies and workflows can drive the definition of RAs by providing structured approaches to requirements specification and providing specific choices of technologies that constrain the design of a reference architecture.

\textbf{Knowledge engineering} From the earliest days of the expert systems era there was a recognition that KE needed a principled methodology~\cite{hayes1983building}, but the first complete realization of such a methodology came in the 1990s with the development of KADS~\cite{wielinga1992kads} and subsequently CommonKADS~\cite{schreiber2000knowledge}. CommonKADS is a methodology for the extraction of expert knowledge into knowledge bases based on lifecycle and corresponding models. CommonKADS has been applied to a variety of domains, from e-governance~\cite{yang2006applying} to multi-agent scenarios~\cite{iglesias1998analysis}. The models formalized by CommonKADS are complemented by MIKE's~\cite{angele1998developing} formalization of the execution of the model, and the Protege~\cite{gennari2003evolution} software for collaborative knowledge production and maintenance. The primary focus of this work is on aspects of task selection, knowledge modeling, and knowledge elicitation, and relatively little attention was paid to architectural aspects and deployment in modern Web-based applications and services, except for the linked data community's emphasis on the use of W3C linked data stack and standards~\cite{hendler2009tonight}. More recently, the growth of Semantic Web applications has resulted in research into semantic patterns~\cite{gangemi2009ontology} and boxologies that organize systems using abstract components~\cite{van2019boxology}. While these boxologies originally aimed to capture purely automated processes, there have been attempts to include human agents, either as process initiators~\cite{van2021modular} or following the human-in-the-loop paradigm~\cite{witschel2020visualization}. With the emergence of knowledge graphs, recent work has devised corresponding workflows for particular domains like the Library and Information Studies (LIS)~\cite{tharani2021much} community, e-commerce applications like the Amazon Product KG~\cite{zalmout2021all}, and generic workflows for the biomedical domain~\cite{lobentanzer2023democratizing}. Finally, there have been attempts to identify common patterns in knowledge graph workflows~\cite{tamavsauskaite2022defining} and design toolkits~\cite{ilievski2020kgtk} that implement these patterns as reusable pipelines of commands. 

\textbf{Ontology engineering} A specific area of focus within knowledge engineering is ontology engineering (OE)~\cite{gomez2006ontological}. The Semantic Web era is characterized by a strong focus on the manual development of ontologies~\cite{noy2001ontology} and their publishing on the Web using linked data principles, with a strong focus on interoperability, reuse, and integration~\cite{poveda2022lot}. Methodologies for ontology engineering developed over the past thirty years include METHONTOLOGY~\cite{fernandez1997methontology}, Kendall and McGuinness's Ontology Development 101~\cite{kendall2019ontology}, and NeOn~\cite{suarez2011neon}. As with the KE methodologies described in the previous section, OE methodologies are concerned with the organizational structures and workflows associated with ontology design, knowledge representation (e.g. the modeling of spatio-temporal modeling~\cite{guillem2023rcc8,ermolayev2014ontologies}), and ontology matching \cite{otero2015ontology}. There has been limited work in understanding the relationship between data governance~\cite{khatri2010designing} and ontology engineering; while both disciplines overlap in their concerns for structuring and managing data, the integration of data governance principles into ontology engineering workflows remains a less explored area. OE and KE workflows are abstracted through the neurosymbolic boxology patterns~\cite{sabouknowledge}, which enables them to be represented in an overarching architecture that is composed of those patterns.

\textbf{Data engineering} Data engineering (DE) itself has provided a wide range of best practices and workflows in common use across the industry. Standard architectural models of data processing systems include data warehouses~\cite{chaudhuri1997overview}, data lakes~\cite{ravat2019data}, and distributed data processing platforms such as Apache Spark~\cite{salloum2016big}. Recent work on \textit{DataOps}~\cite{ereth2018dataops} as an adaptation of DevOps principles and best practices to the design and operation of data processing workflows has established many concepts towards the end of ensuring that data ingestion and integration are smooth, continuous, and error-free. These principles include the monitoring of the quality of data to prevent poor quality or inconsistent data from compromising data integrity of the knowledge graph; data versioning, supporting the ability to revert to previous states of the data or understand changes over time; and designing workflows such that the system can scale accordingly without a compromise in performance \cite{ataei2022state, steidl2023pipeline}. All of these techniques can inform the design of architectures for knowledge engineering. The Andreessen Horowitz reference architecture~\cite{bornstein2020emerging} for emerging data infrastructure and platforms is a snapshot of the current industry stack and trends that subsume most current uses of data within an enterprise. This architecture includes several high-level elements, such as sources, ingestion and transport, storage, query and processing, transformation, and analysis and output. It is noteworthy that, while this architecture has been adapted for artificial intelligence and machine learning workflows, it does not refer at all to knowledge graphs or Semantic Web concepts or products, especially given the care it takes to address specific use cases related to machine learning.

\section{A six-step roadmap to an RA for KE}
\label{sec:roadmap}



By using a requirements-driven approach~\cite{bass2022software,angelov2009classification,cloutier2010concept}, RA methodologies, informed by recent work on DE, KE, and OE methodologies and workflows, can support the consolidation of different perspectives and paradigms under a single umbrella (\textit{challenge 1}). RAs provide a suitable approach for technological alignment by first identifying, consolidating, and prioritizing user needs, followed by formalizing these needs into functional requirements and quality attributes, and, finally, following an iterative development and evaluation of architectures that satisfy these requirements best.
Addressing \textit{challenge 2}, using a requirements-driven iterative design, RAs are developed to suit current technological trends and to be dynamically adapted in the future when the underlying requirements shift significantly. In other words, RAs are designed to be representative of the current technological trends and are flexible to be enhanced over time to suit further developments that are likely to occur in a dynamic field such as KE.
As mainstream software engineering artifacts, RAs can facilitate smoother adoption of KE by software engineers and computer/data scientists (\textit{challenge 3}). An RA is a mechanism for meeting practitioners in such fields halfway and enabling a bridge for seamless integration and collaboration between these fields and KE.

\begin{figure}[!ht]
    \centering
    \includegraphics[width=\linewidth, trim={0 2cm 0 3.5cm},clip]{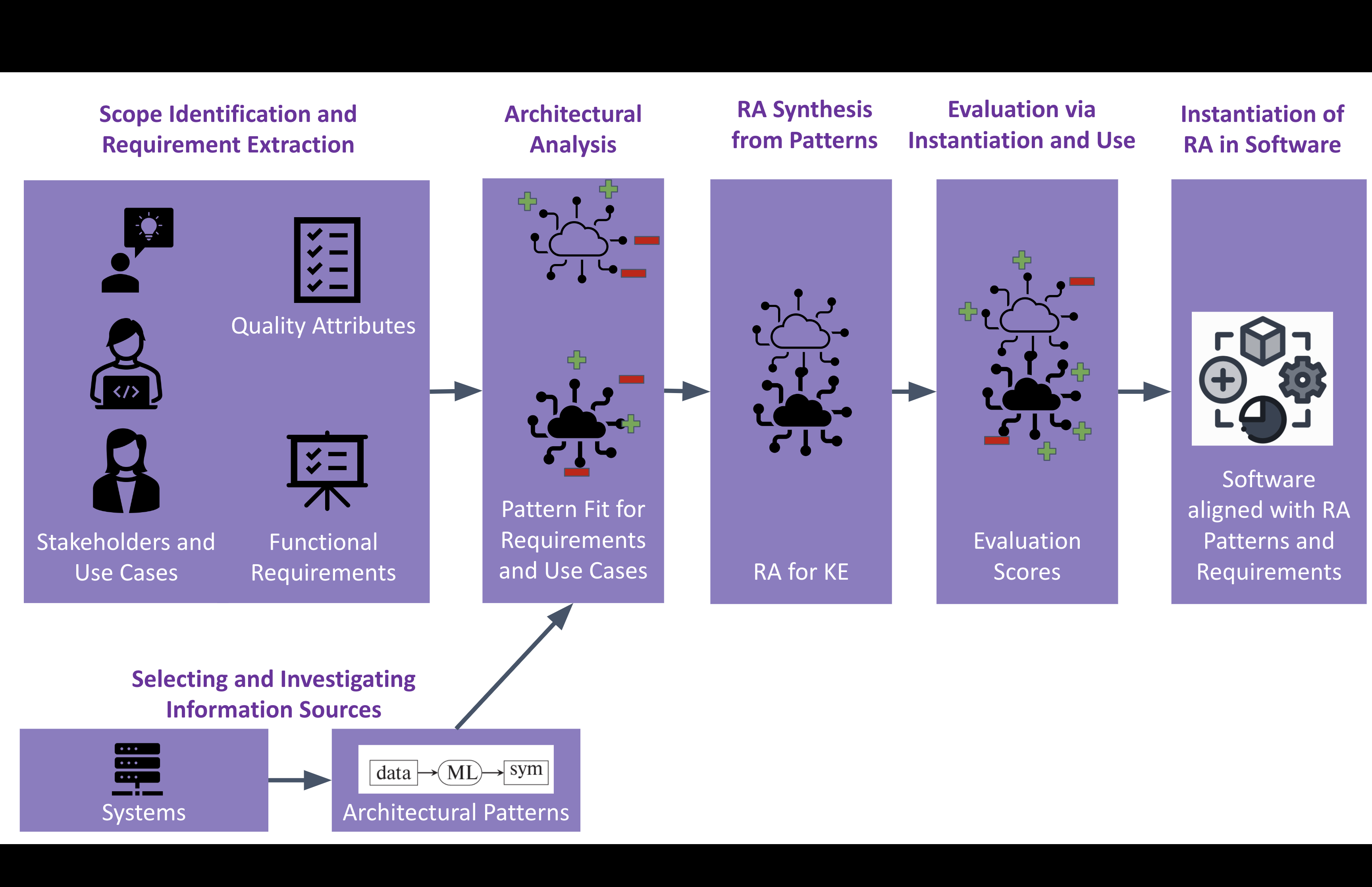}
    \caption{Pipeline for devising an RA for KE. First, we identify the scope by defining stakeholders and use cases, ultimately resulting in a set of quality attributes and functional requirements~\cite{allen2023identifying}. Second, we select and investigate information sources, according to the SWeMLS corpus of neurosymbolic systems and patterns for KE~\cite{ekaputra2023describing,sabouknowledge}. Third, we connect these components through architectural analysis, yielding information about the fit of various patterns for requirements and use cases. Based on these insights, the fourth step synthesizes an RA from these patterns. Fifth, the RA is evaluated through instantiation and use using a standard software architecture methodology. Finally, the RA is instantiated into software.}
    \label{fig:ra-pipeline}
\end{figure}

We propose that RAs for KE should be designed by applying the mainstream software engineering techniques described in the previous section. 
We adopt the methodology proposed by Nakagawa et al.~\cite{nakagawa2011aspect}, consisting of five steps: scope identification (including extraction of requirements), selecting and investigating information sources, architectural analysis, synthesizing an RA, and evaluating the RA through instantiation and use. We include an additional step of instantiating the RA in software, resulting in a six-step procedure. We see the last three steps as iterative steps, which can be modified given
the shifting stakeholder requirements, the modular design of the architecture, and the dynamic nature of the underlying technology for the software implementation. The methodology for devising an RA for KE is summarized in \autoref{fig:ra-pipeline}.

\subsection{Scope identification and extraction of requirements}

We take inspiration from the software engineering practice of using reference architectures as consolidation mechanisms. 
On the one hand, the RA framework needs to cover the \textbf{use cases} that fall under the task of KE. According to our definition and following \cite{sabouknowledge}, KE is a knowledge process that includes knowledge creation (e.g., ontology creation, data ingest) and refinement (e.g., ontology refinement, knowledge graph completion, anomaly detection).
The scope of the RA framework should enable machine, human, and joint machine-human knowledge processes~\cite{tiddi2023knowledge}. The set of tasks that fall within the scope of KE are listed in \autoref{tab:ke_scenarios}, together with a typical scenario and a question that an RA should be designed to solve. For example, the knowledge graph refinement task can be illustrated with a system flagging a potential inconsistency or contradiction. A question for an RA is how it can facilitate the resolution of such quality challenges.

On the other hand, the RA framework must identify and support the \textbf{requirements} of the relevant stakeholders.
In software architecture development~\cite{cloutier2010concept,taylor2010software,bass2022software}, requirements serve as a common denominator to align the needs of the stakeholders and the technical patterns.
While stakeholders may include both knowledge engineers and beneficiaries of KE (e.g., data scientists building applications), we focus on the requirements of knowledge engineers, i.e., users that perform the aforementioned knowledge graph creation and refinement tasks. As is common in software engineering~\cite{bass2022software}, the requirements can be translated into two categories: functional requirements and quality attributes. In recent work~\cite{allen2023identifying}, we devised a set of 23 quality attributes and 8 functional requirements for KE, based on a historiographic analysis of the field of KE (similar to \autoref{ssec:history}). We show an excerpt of five quality attributes and five functional requirements in \autoref{tab:reqs}. These requirements are manually selected to show diverse representative FRs and QAs. Their role is to illustrate to the reader what FRs and QAs for KE (may) look like.
An example of QA is modularity, namely, a requirement that the components of the KE workflow enable selective composition for supporting particular use cases. An example of an FR is the import of common data formats, including mainstream semantic web and software sources, as well as serializations. While we consider~\cite{allen2023identifying} to provide an initial set of FRs and QAs, we note that the review of papers in this prior work is not based on a systematic selection. The work on analyzing Semantic Web and Machine Learning Systems (SWeMLS) identifies three other requirements: maturity, transparency, and auditability, based on a systematically collected set of papers~\cite{sabouknowledge}. Critical future work is to explore how to automatically derive FRs and QAs from such a systematically collected set of papers, based on formally defined requirements. Moreover, given a particular, more narrow scope, the users are expected to define a subset of high-priority requirements that will guide the construction of their RA.





\begin{table*}[!t]
\caption{Example QAs and FRs for KE from~\cite{allen2023identifying}, extended with evaluation criteria. We refer to each requirement with `should' signifying a uniform level of importance. The priority scale of the requirements can be further distinguished according to the specific use case requirements.}
\label{tab:reqs}
\small
\begin{tabular}{p{2cm}|p{4.5cm}|p{7cm}}
\textbf{requirement} & \textbf{description} & \textbf{evaluation criteria} \\ \hline
interoperability \cite{feigenbaum1992personal} & the knowledge produced by the KE process should be easy to share across sites and applications & compatibility with different data formats and standards; ease of integration with other systems; number of supported interfaces/APIs \\
curatability \cite{bender2021dangers} & the KE process should support human curation of automatically extracted and/or inferred knowledge & effectiveness of human curation interfaces; balance between automation and human oversight; quality control measures for curated knowledge
\\
scalability \cite{feigenbaum1992personal} & the KE process should scale economically with the amount of knowledge produced (measured in terms of rules, triples, nodes, edges, etc.) & performance under increasing amounts of knowledge (e.g., response times, throughput);
cost-effectiveness at different scales; system behavior under concurrent user loads \\
modularity \cite{ilievski2020kgtk} & the components of the knowledge engineering process should be selectively composable to suit a specific use case & independence and interchangeability of system components; ability to integrate or detach modules based on need; impact of module changes on overall system performance \\
customizability \cite{ilievski2020kgtk} & the components of the KE process should be modifiable to support specific use cases & ease and extent of system modifications; number of customizable components; user feedback on customization features \\ \hline
supports semantic web standards \cite{berners2001semantic} & the KE process should support the use of W3C semantic web standards & use of standard knowledge representation (e.g., RDF, property graphs), serializations (e.g. Turtle, JSON-LD) and query languages (e.g., SPARQL, Cypher), evaluated by ontology quality metrics, pitfall scanning \\
imports common data formats \cite{ilievski2020kgtk} & the KE process should support the import of data and/or knowledge from data sources & use of standard serializations (e.g., CSV, JSON, Parquet), evaluated by ontology quality metrics, parsing error rate\\
exports common data formats \cite{ilievski2020kgtk} & the produced knowledge should be exportable to software industry-standard data delivery mechanisms  & use of software industry-standard data storage mechanisms (e.g., relational databases, RDF data dumps, search engine indexes) and integration standards (e.g., serialized data dumps, publish/subscribe messaging, REST APIs), evaluated by time to deploy, storage and compute costs\\
provides user-friendly interfaces \cite{ilievski2020kgtk} & the knowledge produced by the KE process should be accessible and applicable by end users & industry-standard user experience (e.g., command line interfaces, visual editors and browsers, reporting and analytics dashboards) measured by time to complete tasks, user satisfaction surveys \\
supports heterogeneous query \cite{hartig2022reflections} & the knowledge produced by the KE process should be searchable using multiple query languages & use of industry-standard query languages  (e.g., SQL, Cypher, SPARQL) and query execution strategies (e.g., federated query, centralized query, find-and-follow), with developer experience measured by time to complete tasks, user satisfaction surveys \\
\hline
\end{tabular}%
\end{table*}

\subsection{Selection and investigation of information sources}

\begin{figure}[!ht]
    \centering
    \includegraphics[width=\linewidth]{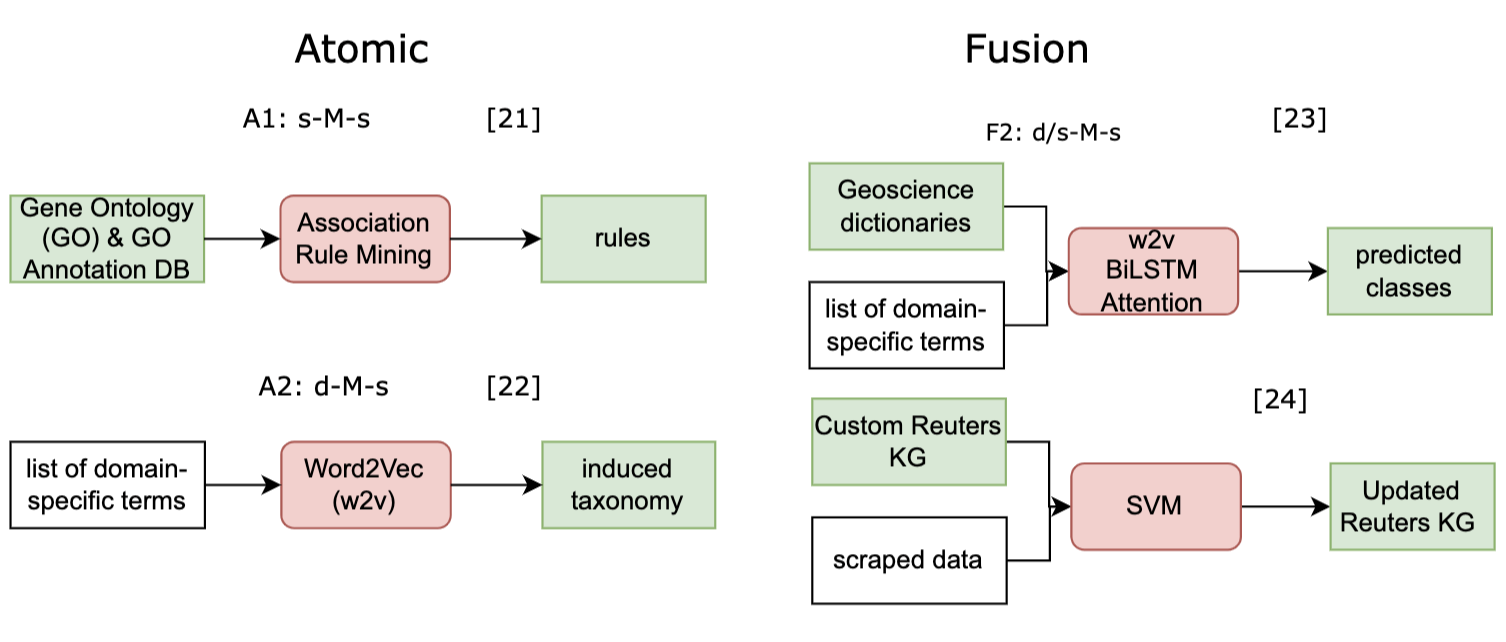}
    \caption{Simple neurosymbolic system design patterns from the SWeMLS KG, as shown in \cite{waltersdorfer2023semantic}. The F2 design pattern, appearing on the right of the figure, is a simple fusion that takes both symbolic (s) and unstructured data (d) as inputs and produces symbolic data (s) as output using a model M.}
    \label{fig:simple-swemls-patterns}
\end{figure}

A systematic analysis of the NeSy landscape, aiming to characterize SWeMLS published between 2010 and 2020, resulted in a corpus of 476 system papers~\cite{breit2023combining}. In this work, each of the papers was annotated with bibliographic information (authors, institutions, publication year, and venue), domain of application, task solved, input/output system architecture, characteristics of the machine learning and the semantic web modules, and levels of maturity, transparency, and provenance. The system components are aligned to the boxology for neurosymbolic systems~\cite{van2019boxology}. In total, 44 patterns were discovered, classified into a typology of six types according to their shapes.
Some example boxology patterns from the SWeMLS corpus are shown in \autoref{fig:simple-swemls-patterns}. The F2 pattern (short for \textit{fusion-2}) is described in \cite{waltersdorfer2023semantic} as a simple fusion design pattern that takes both symbolic (s) and unstructured data (d) as inputs and produces symbolic data (s) as output using a model M. The F2 pattern corresponds to two specific systems, one being a geological text document classifier \cite{qiu2020dictionary}, and the other an application that classifies heterogeneous web content to create symbolic data extending an enterprise knowledge graph \cite{song2017building}.

The data from the study by \cite{ekaputra2023describing} is made available as a knowledge graph. The ontology of this knowledge graph is centered around the class \texttt{System}, which belongs to one \texttt{Pattern} and has $N$ \texttt{System Component} values. Using SPARQL queries against the SWeMLS knowledge graph, we identified a subset of 139 papers as KE-related, consisting of papers whose systems perform \texttt{Graph creation} or \texttt{Graph extension} tasks, and produce \texttt{Symbol} as the final output. In doing so, we followed the procedure described in \cite{sabouknowledge}.
We use this set of 139 KE papers in the rest of our methodology, given the systematic approach to collecting them, their rich annotation, and their alignment with the NeSy boxology components. This procedure illustrates how the selection of information sources can be achieved - in practice, RA developers may decide to focus on a different set of sources, e.g., covering a larger set of papers or a particular subarea of KE for better representativeness to their envisioned use cases.


\begin{figure}[!ht]
    \centering
    \includegraphics[width=\linewidth]{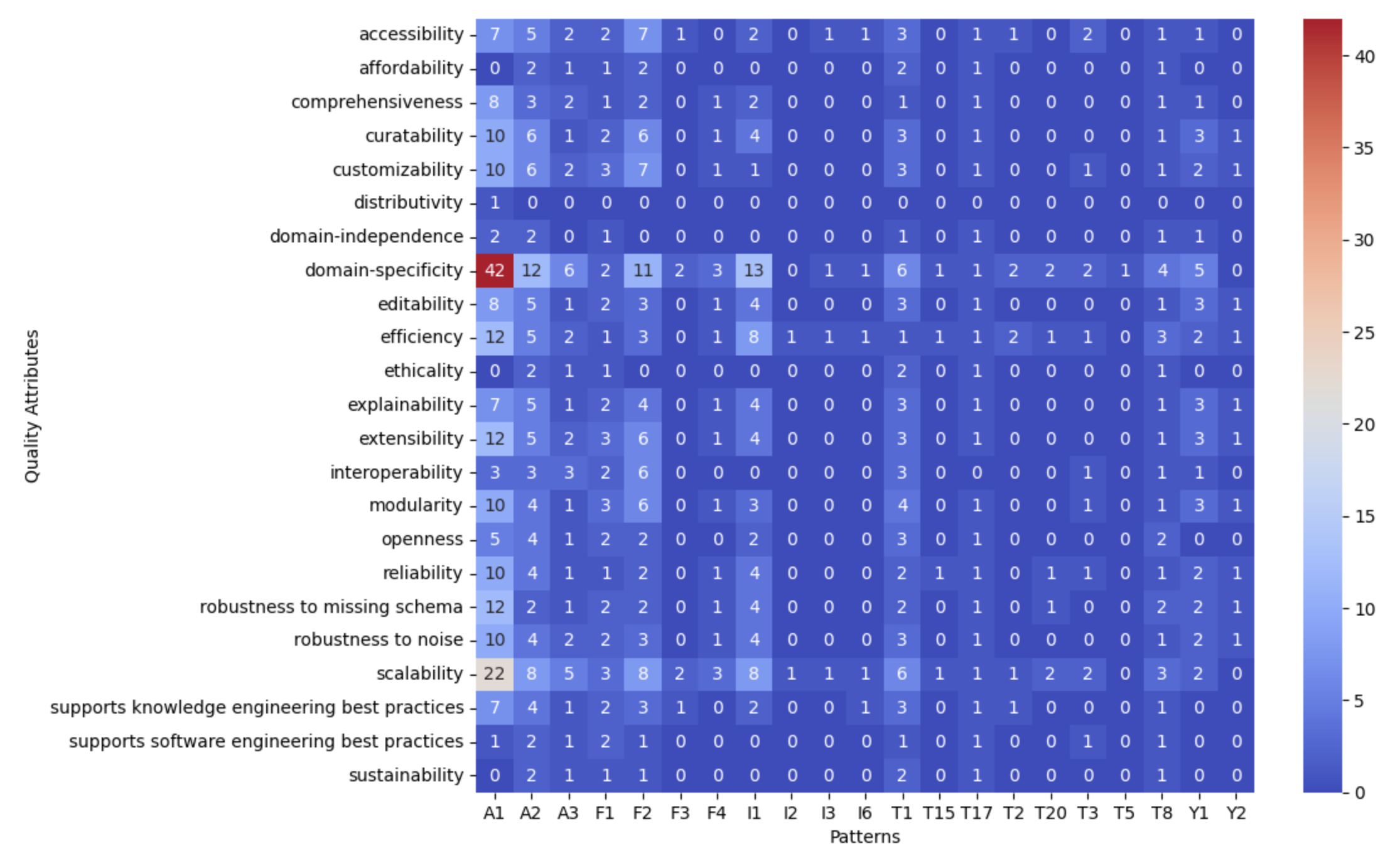}
    \caption{Preliminary analysis of the relationships between quality attributes for KE identified in \cite{allen2023identifying} and the KE design patterns from \cite{sabouknowledge} that are associated with knowledge graph creation and extension. The number in each cell is the count of occurrences of the quality attributes assigned to papers by the zero-shot text classifier that describes systems with the given pattern.}
    \label{fig:qa_x_pattern}
\end{figure}

\subsection{Architectural analysis}

The next step is to perform a preliminary analysis of the extent to which quality attributes for KE are supported within specific SWeMLS patterns. We illustrate this analysis over the SWeMLS KG, where papers describe a system, and each system is associated with a specific pattern. To establish a connection between quality attributes and patterns, we utilize the SerpApi Google Scholar API to obtain snippets from the abstracts of each of the 139 papers described in the previous section.\footnote{\url{https://serpapi.com/google-scholar-api}, accessed: 2024-01-05.} We then construct a zero-shot text classifier using prompt programming of ChatGPT \cite{schade2023how} that, given an article's snippet and title, assigns one or more quality attributes to each paper. Here, we used the 23 QAs identified in the first step. We then aggregate the quality attributes for each paper's system's pattern across all of the papers and patterns. This allows us to derive a \textbf{matrix relating quality attributes to patterns}, as shown in Figure \ref{fig:qa_x_pattern}. From this initial analysis, we find that the A1, A2, F2, and T1 patterns cover the most quality attributes. Namely, A1 covers 20 of the 23 QAs, except for affordability, ethicality, and sustainability; A2 and T1 only lack the QA of distributivity; and F2 covers 20 QAs lacking only distributivity, domain independence, and ethicality. We find these insights to be largely intuitive, as many systems belong to patterns such as A1 and F2. Among the quality attributes, we note that most patterns capture domain-specificity and scalability, whereas ethicality and distributivity are rare. This indicates the tendency of neurosymbolic KE systems to focus on scalability and domain-specificity, whereas aspects such as sustainability, ethicality, and distributivity are gradually gaining momentum but are not yet a primary consideration for most systems.

We emphasize that the corpus used for our analysis is not comprehensive and that the specific analytical methodology followed in this paper may exhibit classification bias. Thus, the significance of this analysis is mainly to show an illustration of how architectural patterns and quality attributes can be linked together. This provides us with a means to determine, given the quality attributes and functional requirements from the scope identification and requirements extraction steps, which pattern(s) are candidates for RA synthesis. We leave it to future work to further tune this procedure, generalize it to a larger dataset, and devise a more robust classification engine. Finally, we note that an analogous procedure can be followed for aligning functional requirements with boxology patterns.

\subsection{RA synthesis from patterns}

\subsubsection{Procedure}

Given the architectural analysis of the patterns from the boxology and from other prominent workflows, the construction of the RA follows as a natural synthesis step. Namely, the RA consolidates the discovered pattern(s) with consideration for their adequacy for addressing the use cases and the derived requirements. The benefit of this synthesis is that it prescribes a global view of how a given pattern addressed the stakeholder needs, how the different patterns fit together if a complex pattern is being composed of simpler patterns, and how the architectural pattern(s) can be technically realized; all of that, while aligning with state-of-the-art workflows as reported in the literature.  


\begin{figure}[!ht]
    \centering
    \includegraphics[width=\linewidth]{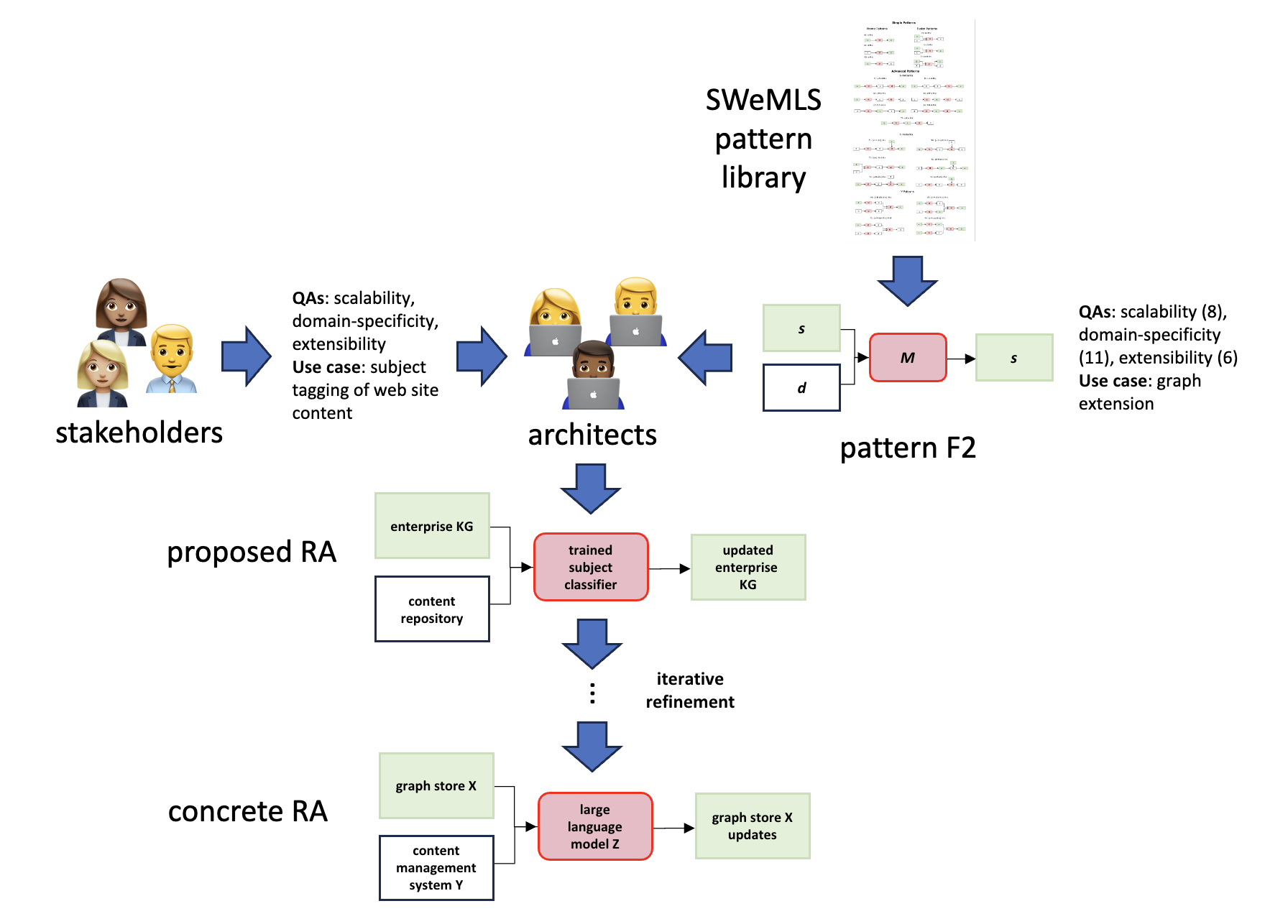}
    \caption{An example of RA synthesis. Stakeholders have identified a set of QAs (scalability, domain-specificity, and extensibility) and a specific use case (graph extension of an enterprise KG).  The team of architects has taken this as input, selected an adequate pattern F2 (fusion 2) based on its support for the indicated QAs and use case, and synthesized a proposed RA that uses a trained subject classifier to perform graph extension based on the KG and data from a content repository. After a process of iterative refinement, choices are made about specific technologies to use, and a concrete RA is proposed.}
    \label{fig:RA-synthesis-from-patterns}
\end{figure}

How would this synthesis of patterns into an RA be realized in practice?
As the synthesis is highly dependent on the high-priority requirements of the RA stakeholders, it is impossible to prescribe a one-size-fits-all architecture. Instead, we describe the procedure of how an RA would be synthesized for a specific use case, illustrated in Figure \ref{fig:RA-synthesis-from-patterns}.
With the prioritized QAs (from Step 1) as a guide, components (from Step 2) that address these attributes (using the analysis from Step 3) will be identified and integrated into the architecture. Practically, an initial design meeting would be scheduled to review the existing workflows and patterns concerning the requirements. During this meeting, a candidate RA will be crafted, following high-level architectural principles and approaches.  A core team of architects is then essential to conduct a collaborative design session. In this session, the RA's architecture, its components, and their interactions are laid out, creating a platform for real-time discussions and potential modifications. During these discussions, the SWeMLS knowledge graph can provide information about potential alternative technologies for the components. To refine the design further, a series of subsequent sessions can be organized that invite a broader set of participants. The feedback gathered from these sessions can be used to iterate and enhance the design. 

Following a similar procedure, an example RA for KE in the domain of newsrooms and journalism has been provided by \cite{ocana2023software}. A critical future work is to apply our process to other use cases with potentially different requirements.

\subsubsection{Hypothetical scenario} We proceed to illustrate this process with a hypothetical \textbf{scenario} (Figure \ref{fig:RA-synthesis-from-patterns}). 
Business stockholders at a large enterprise have identified that there is a need to improve the discoverability of content on a corporate website. The company has a repository of content, including product information, articles, blog posts, case studies, and user guides. However, users often struggle to find the content they need, leading to frustration, reduced engagement, and potentially lost sales opportunities. The company recognizes that better content recommendations and more intuitive navigation can significantly enhance the user experience and drive business outcomes.

Based on their understanding of industry best practices, the stakeholders determine that a solution that performs subject tagging of content using a domain-specific ontology will support the discoverability of content for tasks their users are attempting to accomplish. They identify several \textbf{quality attributes} that they want to ensure such a solution addresses:
\begin{itemize}
\item Scalability: The solution should handle a large and growing volume of content and user interactions.
\item Domain-specificity: The solution should provide subject tagging from a domain-specific taxonomy of vocabulary terms and definitions.
\item Extensibility: The solution should be extensible as new content and subjects become available.
\end{itemize}

These QAs, along with additional FRs, are presented to a team of architects. The architects review the \textbf{design patterns} captured in the SWeMLS KG to determine which of the identified design patterns most closely address these QAs and requirements. The knowledge graph supports the review process by surfacing relevant papers, case studies, and benchmarks for similar systems and use cases. Querying the KG, the architects identify the F2 design pattern as describing systems that match the stakeholder QAs and the use case. Both of the systems corresponding to F2 are similar to the stakeholder use case, and both provide evidence that the pattern can address the specified QAs. Based on this review, the \textbf{F2} design pattern is selected.

The architects then specify how the requirements and use case can be addressed by instantiating the components in the F2 design pattern into a \textbf{proposed RA}, as follows:
\begin{itemize}
\item The input symbolic representation (s) is an enterprise KG that captures the semantics of the content repository and application domain, including the domain subject taxonomy. The KG should capture key entities like products, articles, and customer segments, along with their relationships.
\item The input data (d) is the content on the corporate website.
\item The model (M) component is a combination of ML and NLP technologies that given the KG and content, classifies the content according to the domain-specific subject taxonomy.
\item The output symbolic representation (s) are relations to be added to the knowledge graph to link content on the website to relevant concepts in the vocabulary.
\end{itemize} 

Given these decisions, the architects then proceed to make additional choices for what specific technologies are to be used to implement each component into a \textbf{concrete RA}:
\begin{itemize}
\item Knowledge graph (s): this could be stored in a labeled property graph database, an RDF triple store, or a relational database with a graph-friendly schema. 
\item Content repository  (d): given the existing website, the input data may reside in various enterprise systems like content management systems or customer-customer relationship management systems.
\item Model (M): the model is responsible for producing multi-class subject classifications from the input data and KG, and updating the KG with this new knowledge. Suitable approaches could include 
hybrid models that combine text, user interactions, and graph structure, e.g., using transformer architectures like BERT or pre-trained language models like GPT-4.
\end{itemize}

The architects additionally consider factors such as the volume and variety of input data, the complexity of the topic taxonomy, explainability requirements, and the team's AI/ML skills in making their decisions about how to instantiate the RA, including the following considerations:
\begin{itemize}
\item The iterative nature of the F2 pattern, where the output enhances the KG, can support continuous improvement of recommendations as new content is incorporated.
\item The use of a KG as the core representation to aid explainability, as the relationships between content and topics can be traced and visualized, potentially helping content managers optimize the content strategy and troubleshoot issues.
\item The separation of concerns in the F2 design pattern, with dedicated components for data ingestion, model training, and KG management, promotes scalability and performance, as each component can be independently optimized and scaled based on the workload.
\end{itemize}

The architects then go through a final process of determining the \textbf{final proposed architecture}, potentially including:
\begin{itemize}
\item Assessing the current state of the KG and identifying gaps in topic coverage
\item Inventorying available data sources and evaluating their relevance and quality
\item Experimenting with different modeling approaches and comparing their accuracy, scalability, and interpretability
\item Validate model outputs with subject matter experts and through user testing
\end{itemize}

The final proposed architecture is then documented to support the evaluation process described in the next phase of the process.
Once the reference architecture has been defined, it can be \textbf{stored} in the SWeMLS KG. This allows the RA to be shared and reused in other content classification applications within the enterprise. Some examples of how elements of the reference architecture definition can be mapped into the SWeMLS knowledge graph are:
\begin{itemize}
\item The overall RA for content recommendation can be represented as an instance of the \texttt{swemls:System} class. The specific pattern it implements (F2) can be indicated using the \texttt{swemls:hasCorrespondingPattern} property.
\item The business problem of improving content discoverability on the corporate website can be described using the \texttt{swemls:Task} class.
\item The various data sources used to build and enhance the KG, such as content metadata, user interaction logs, and external taxonomies, can be captured using the \texttt{swemls:Data} class.
\item The KG serving as the core symbolic representation can be modeled as an instance of the \texttt{swemls:SemanticWebResource} class. The specific KG technology used can be specified using the \texttt{swTechnology} property.
\item The machine learning model used to learn topic classifications can be represented using the \texttt{swemls:Model} class.
\item The process of training the machine learning models using the input data and KG can be represented using the \texttt{swemls:ProcessingEngine} class.
\item The specific tools, libraries, and frameworks used to implement the RA components can be captured as instances of the relevant classes, including \texttt{swemls:Data}, \texttt{swemls:Model}, and \texttt{swemls:SemanticWebResource}.
\end{itemize}

By mapping the RA to the SWeMLS ontology in this way, we establish a structured and semantically rich representation of the architectural knowledge that can be a resource in the evaluation process described in the next section. The ontology classes, properties, and relationships provide a standardized vocabulary to describe the various aspects of the RA, from the business goals and QAs to the technical components and best practices. This consistent representation facilitates comparison, integration, and reasoning across different RAs and domain applications.

\subsection{RA evaluation through instantiation and use}

Once specified, architectures synthesized in this manner from design patterns can then be evaluated through a lightweight version of the Architecture Tradeoff Analysis Method (ATAM). 
ATAM \cite{kazman2000atam} is a risk-mitigation process used to identify architectural risks that have implications in fulfilling quality attributes. As originally proposed by CMU's Software Engineering Institute, this process involved a multi-day face-to-face gathering of stakeholders and architects. A lightweight ATAM process \cite{sahlabadi2022lightweight} is a streamlined version of the traditional ATAM, focusing on a shorter, more rapid timeframe and often less resource-intensive evaluation of architectural decisions. The use of web-conferencing and real-time collaborative document editors allows this process to be conducted remotely, increasing the ability to gather a large and diverse group of stakeholders. First, we will identify and recruit a set of stakeholders for an ATAM session. On the day of the session, the stakeholders will be presented with an agenda for the session that defines the scope of the evaluation and presents the RA, identifying architectural approaches used. This will be followed by a presentation of user scenarios relative to evaluating the RA. The user scenarios for knowledge creation and maintenance processes should include capabilities for data integration from multiple structured sources \cite{ekaputra2017ontology}, data quality checks \cite{preece2001evaluating}, entity resolution \cite{christophides2020overview}, ontology merging and alignment \cite{chatterjee2018ontology, otero2015ontology}, query optimization \cite{hartig2009executing}, and natural language processing \cite{schneider2022decade} (cf. \autoref{tab:ke_scenarios}). Moreover, the production processes should be automated to enable efficient updates and maintenance of the knowledge artifact \cite{noy2019industry}. In cases where the KE involves the use of personally identifiable information or other sensitive data for knowledge elicitation or training ML components, there is the danger of leakage of sensitive information; in addition, ML components themselves can inadvertently leak data under adversarial attack~\cite{gallofre2020data}. Therefore, the production process should incorporate mechanisms for security and privacy, as well as access control mechanisms to ensure that the data stays secure and that only authorized users have access. 
It is worth observing that many of these issues have been explored to date in the more generic context of data engineering and data science architectures and platforms. Once the stakeholders have considered the various scenarios, they can proceed to collaboratively analyze the scenarios, identifying risks and trade-offs, and gathering feedback, focusing on potential refinements and architectural alternatives. The stakeholders will then document risks, trade-offs, architectural decisions, and the reasons for them, finishing by summarizing the final consensus RA.

\subsection{RA instantiation in a concrete software architecture}
Given a consensus RA, we can proceed to finalize a comprehensive architectural blueprint. 
The RA does not provide absolute recommendations on such choices, assuming that those are stakeholder need-dependent. It does, however, prescribe an association between different requirements, architectural patterns, and adequate implementations.
For each component, the range of options for its instantiation using existing software packages or through bespoke development will be identified. Here, we are inspired by recent toolkits for knowledge graphs, like KGTK~\cite{ilievski2020kgtk}, which connect knowledge engineering operations by defining a universal interface format and abstracting the implementation of each component from the user. The implementation of each component relies on thorough research and consideration of the best existing tool or implementation that can be wrapped, i.e., that the software can provide an interface to. For instance, one could provide an interface to Pytorch-Biggraph~\cite{lerer2019pytorch} for knowledge completion, Shape expressions (Shex) tools~\cite{thornton2019using} for using constraints to evaluate quality, and RLTK~\cite{yao2019extensible} for record linkage across knowledge artifacts. However, as KGTK and similar toolkits are built based on an implicit set of use cases and user requirements, further investigation is required to assess whether they will align with emerging architectural contributions like the set of boxology patterns.

A strategic approach to instantiating an RA involves a phased implementation. Each phase should predominantly focus on one specific component. During this development phase, constant testing and evaluation of each component will be performed to ensure the component aligns with the predefined QAs and specific scenarios. After the conclusion of each phase, feedback will be gathered from all involved stakeholders. This iterative process will ensure the architecture remains relevant and effective, as necessary revisions based on the feedback can be made. The resulting implementation will be open-sourced and thoroughly documented in a publicly accessible code repository. After the entire process is complete, the system's efficacy will be tested in pilot trials facilitated by the stakeholders. During these trials, relevant data on system performance and quality assessment will be collected to ensure the architecture's robustness and efficiency. Notably, the implemented RA would serve as a comprehensive framework that enables decisions on technology for representation, integration, and quality assurance, among others, to be made based on high-priority requirements. While the implementation is meant to be prescriptive and enable efficient KE by profiles with various backgrounds, goals, and levels of expertise, we acknowledge that these recommendations should be considered relative to the stakeholders' needs. 

\section{Conclusions}
\label{sec:conc}


Knowledge engineering, as a process of creating and maintaining knowledge artifacts, has remained relevant throughout the history of AI. In light of the heterogeneous requirements and KE use cases, on the one hand, and the emergence of architectural components and partial workflows, on the other hand, this paper makes a case for developing reference architectures for KE. Following software engineering practices, an RA would provide an organizational principle for isolated systemic patterns, thus providing a key contribution to this ongoing work that enables the knowledge engineering field to be systematized. A reference architecture consolidates the patterns while simultaneously considering its scope, defined through a set of use cases and their corresponding requirements, distilled as quality attributes and functional requirements. The synthesis of the architecture is an iterative process, inspired by success stories of reference architectures for service-oriented design, e-government, and the automotive sector. A key aspect of the development is its evaluation through instantiation and use with representative users for representative KE tasks. As a final step, the reference architecture components need to be instantiated into software, thus closing the cycle between user needs and existing technological capabilities.

While this paper outlines a roadmap for devising comprehensive and requirement-grounded RAs for KE, its realization in practice is partial at present. We present a broad definition of scope through a definition of representative tasks and distillation of 23 quality attributes and 8 functional requirements, which could be narrowed down given a specific use case. We take the recently identified collection of system patterns for neurosymbolic KE as information sources, providing initial components that can be used to construct an RA. We present an architectural analysis, as a direct mapping between QAs and the identified architectural patterns, detecting requirements with various levels of support. The steps of synthesizing an RA from patterns, evaluating the RA through instantiation and use, and instantiating the RA into software are presented as prescriptive, consolidating best practices and methodologies from software engineering through step-by-step processes, because these steps are highly dependent on the specific use cases. Each of these steps requires the dedicated effort of iterative design, development, implementation, and evaluation of an RA, which we plan to pursue as the next steps for representative subsets of tasks and domains. We believe that the presented methodology for devising RAs for KE provides an important extension of emerging work that systematizes KE methods, by providing a mechanism to associate architectural patterns with user requirements and identify potential gaps. We invite the broader community of interested researchers and developers to join us in these discussions and complement our future efforts in consolidating KE practices.

\bibliography{tgdkbib}

\begin{thebibliography}{10}

\bibitem{abian2022analysis}
David Abi{\'a}n, Albert Mero{\~n}o-Pe{\~n}uela, and Elena Simperl.
\newblock An analysis of content gaps versus user needs in the wikidata knowledge graph.
\newblock In {\em International Semantic Web Conference}, pages 354--374. Springer, 2022.

\bibitem{alkhamissi2022review}
Badr AlKhamissi, Millicent Li, Asli Celikyilmaz, Mona Diab, and Marjan Ghazvininejad.
\newblock A review on language models as knowledge bases.
\newblock {\em arXiv preprint arXiv:2204.06031}, 2022.

\bibitem{allen2023identifying}
Bradley~P. Allen, Filip Ilievski, and Saurav Joshi.
\newblock Identifying and consolidating knowledge engineering requirements.
\newblock {\em arXiv preprint arXiv:2306.15124}, 2023.
\newblock \href {https://arxiv.org/abs/2306.15124} {\path{arXiv:2306.15124}}.

\bibitem{angele1998developing}
J{\"u}rgen Angele, Dieter Fensel, Dieter Landes, and Rudi Studer.
\newblock Developing knowledge-based systems with mike.
\newblock {\em domain modelling for interactive systems design}, pages 9--38, 1998.

\bibitem{angelov2009classification}
Samuil Angelov, Paul Grefen, and Danny Greefhorst.
\newblock A classification of software reference architectures: Analyzing their success and effectiveness.
\newblock In {\em 2009 Joint Working IEEE/IFIP Conference on Software Architecture \& European Conference on Software Architecture}, pages 141--150. IEEE, 2009.

\bibitem{arsanjani2007s3}
Ali Arsanjani, Liang-Jie Zhang, Michael Ellis, Abdul Allam, and Kishore Channabasavaiah.
\newblock S3: A service-oriented reference architecture.
\newblock {\em IT professional}, 9(3):10--17, 2007.

\bibitem{ataei2022state}
Pouya Ataei and Alan Litchfield.
\newblock The state of big data reference architectures: A systematic literature review.
\newblock {\em IEEE Access}, 2022.

\bibitem{bass2022software}
Len Bass, Paul Clements, and Rick Kazman.
\newblock {\em Software architecture in practice}.
\newblock SEI Series in Software Engineering. Addison-Wesley Professional, fourth edition, 2022.

\bibitem{beek2016lod}
Wouter Beek, Laurens Rietveld, Stefan Schlobach, and Frank van Harmelen.
\newblock Lod laundromat: Why the semantic web needs centralization (even if we don't like it).
\newblock {\em IEEE Internet Computing}, 20(2):78--81, 2016.

\bibitem{bender2021dangers}
Emily~M. Bender, Timnit Gebru, Angelina McMillan-Major, and Shmargaret Shmitchell.
\newblock On the dangers of stochastic parrots: Can language models be too big?
\newblock In {\em Proceedings of the 2021 ACM Conference on Fairness, Accountability, and Transparency}, FAccT '21, page 610–623, New York, NY, USA, 2021. Association for Computing Machinery.
\newblock \href {https://doi.org/10.1145/3442188.3445922} {\path{doi:10.1145/3442188.3445922}}.

\bibitem{berners2001semantic}
Tim Berners-Lee, James Hendler, and Ora Lassila.
\newblock The semantic web.
\newblock {\em Scientific american}, 284(5):34--43, 2001.

\bibitem{boldt95}
Juergen Boldt.
\newblock The common object request broker: Architecture and specification.
\newblock Specification formal/97-02-25, Object Management Group, July 1995.
\newblock URL: \url{http://www.omg.org/cgi-bin/doc?formal/97-02-25}.

\bibitem{bornstein2020emerging}
Matt Bornstein, Jennifer Li, and Casado. Martin.
\newblock Emerging architectures for modern data infrastructure.
\newblock \url{https://future.com/emerging-architectures-modern-data-infrastructure/}, 2020.
\newblock Accessed: 2022-12-02.

\bibitem{breit2023combining}
Anna Breit, Laura Waltersdorfer, Fajar~J Ekaputra, Marta Sabou, Andreas Ekelhart, Andreea Iana, Heiko Paulheim, Jan Portisch, Artem Revenko, Annette~ten Teije, et~al.
\newblock Combining machine learning and semantic web: A systematic mapping study.
\newblock {\em ACM Computing Surveys}, 2023.

\bibitem{chatterjee2018ontology}
Niladri Chatterjee, Neha Kaushik, Deepali Gupta, and Ramneek Bhatia.
\newblock Ontology merging: A practical perspective.
\newblock In {\em Information and Communication Technology for Intelligent Systems (ICTIS 2017)-Volume 2 2}, pages 136--145. Springer, 2018.

\bibitem{chaudhuri1997overview}
Surajit Chaudhuri and Umeshwar Dayal.
\newblock An overview of data warehousing and olap technology.
\newblock {\em ACM Sigmod record}, 26(1):65--74, 1997.

\bibitem{christophides2020overview}
Vassilis Christophides, Vasilis Efthymiou, Themis Palpanas, George Papadakis, and Kostas Stefanidis.
\newblock An overview of end-to-end entity resolution for big data.
\newblock {\em ACM Computing Surveys (CSUR)}, 53(6):1--42, 2020.

\bibitem{cloutier2010concept}
Robert Cloutier, Gerrit Muller, Dinesh Verma, Roshanak Nilchiani, Eirik Hole, and Mary Bone.
\newblock The concept of reference architectures.
\newblock {\em Systems Engineering}, 13(1):14--27, 2010.

\bibitem{dong2015schema}
Xin~Luna Dong and Divesh Srivastava.
\newblock Schema alignment.
\newblock In {\em Big Data Integration}, pages 31--61. Springer, 2015.

\bibitem{ekaputra2017ontology}
Fajar Ekaputra, Marta Sabou, Estefan{\'\i}a Serral~Asensio, Elmar Kiesling, and Stefan Biffl.
\newblock Ontology-based data integration in multi-disciplinary engineering environments: A review.
\newblock {\em Open Journal of Information Systems}, 4(1):1--26, 2017.

\bibitem{ekaputra2023describing}
Fajar~J Ekaputra, Majlinda Llugiqi, Marta Sabou, Andreas Ekelhart, Heiko Paulheim, Anna Breit, Artem Revenko, Laura Waltersdorfer, Kheir~Eddine Farfar, and S{\"o}ren Auer.
\newblock Describing and organizing semantic web and machine learning systems in the swemls-kg.
\newblock In {\em European Semantic Web Conference}, pages 372--389. Springer, 2023.

\bibitem{ereth2018dataops}
Julian Ereth.
\newblock Dataops-towards a definition.
\newblock {\em LWDA}, 2191:104--112, 2018.

\bibitem{ermolayev2014ontologies}
Vadim Ermolayev, Sotiris Batsakis, Natalya Keberle, Olga Tatarintseva, and Grigoris Antoniou.
\newblock Ontologies of time: Review and trends.
\newblock {\em International Journal of Computer Science \& Applications}, 11(3), 2014.

\bibitem{feigenbaum1977art}
Edward~A Feigenbaum.
\newblock The art of artificial intelligence: Themes and case studies of knowledge engineering.
\newblock In {\em Proceedings of the Fifth International Joint Conference on Artificial Intelligence}, volume~2. Boston, 1977.

\bibitem{feigenbaum1992personal}
Edward~A. Feigenbaum.
\newblock A personal view of expert systems: Looking back and looking ahead.
\newblock {\em Expert Systems with Applications}, 5(3):193--201, 1992.
\newblock Special Issue: The World Congress on Expert System.
\newblock URL: \url{https://www.sciencedirect.com/science/article/pii/095741749290004C}, \href {https://doi.org/10.1016/0957-4174(92)90004-C} {\path{doi:10.1016/0957-4174(92)90004-C}}.

\bibitem{fernandez1997methontology}
Mariano Fernández-López, Asuncion Gomez-Perez, and Natalia Juristo.
\newblock Methontology: from ontological art towards ontological engineering.
\newblock {\em Engineering Workshop on Ontological Engineering (AAAI97)}, 03 1997.

\bibitem{framework2015draft}
DRAFT NIST Big Data~Interoperability Framework.
\newblock Draft nist big data interoperability framework: Volume 6, reference architecture.
\newblock {\em NIST Special Publication}, 1500:6, 2015.

\bibitem{gangemi2009ontology}
Aldo Gangemi and Valentina Presutti.
\newblock Ontology design patterns.
\newblock In {\em Handbook on ontologies}, pages 221--243. Springer, 2009.

\bibitem{garces2021three}
Lina Garc{\'e}s, Silverio Mart{\'\i}nez-Fern{\'a}ndez, Lucas Oliveira, Pedro Valle, Claudia Ayala, Xavier Franch, and Elisa~Yumi Nakagawa.
\newblock Three decades of software reference architectures: A systematic mapping study.
\newblock {\em Journal of Systems and Software}, 179:111004, 2021.

\bibitem{gennari2003evolution}
John~H Gennari, Mark~A Musen, Ray~W Fergerson, William~E Grosso, Monica Crub{\'e}zy, Henrik Eriksson, Natalya~F Noy, and Samson~W Tu.
\newblock The evolution of prot{\'e}g{\'e}: an environment for knowledge-based systems development.
\newblock {\em International Journal of Human-computer studies}, 58(1):89--123, 2003.

\bibitem{goebel2008role}
Randy Goebel, Sandra Zilles, Christoph Ringlstetter, Andreas Dengel, and Gunnar~Aastrand Grimnes.
\newblock What is the role of the semantic layer cake for guiding the use of knowledge representation and machine learning in the development of the semantic web?
\newblock In {\em AAAI Spring Symposium: Symbiotic Relationships between Semantic Web and Knowledge Engineering}, pages 45--50, 2008.

\bibitem{gomez2006ontological}
Asunci{\'o}n G{\'o}mez-P{\'e}rez, Mariano Fern{\'a}ndez-L{\'o}pez, and Oscar Corcho.
\newblock {\em Ontological Engineering: with examples from the areas of Knowledge Management, e-Commerce and the Semantic Web}.
\newblock Springer Science \& Business Media, 2006.

\bibitem{groth2023knowledge}
Paul Groth, Elena Simperl, Marieke van Erp, and Denny Vrande{\v{c}}i{\'c}.
\newblock Knowledge graphs and their role in the knowledge engineering of the 21st century (dagstuhl seminar 22372).
\newblock {\em Dagstuhl Reports}, 12(9), 2023.

\bibitem{guan2023leveraging}
Lin Guan, Karthik Valmeekam, Sarath Sreedharan, and Subbarao Kambhampati.
\newblock Leveraging pre-trained large language models to construct and utilize world models for model-based task planning.
\newblock {\em Advances in Neural Information Processing Systems}, 36:79081--79094, 2023.

\bibitem{guillem2023rcc8}
Ana{\"\i}s Guillem, Antoine Gros, K{\'e}vin R{\'e}by, Violette Abergel, and Livio De~Luca.
\newblock Rcc8 for cidoc crm: semantic modeling of mereological and topological spatial relations in notre-dame de paris.
\newblock In {\em SWODCH’23: International Workshop on Semantic Web and Ontology Design for Cultural Heritage}, 2023.

\bibitem{hartig2022reflections}
Olaf Hartig.
\newblock {Reflections on Linked Data Querying and other Related Topics}.
\newblock \url{https://olafhartig.de/slides/Slides-DKG-SWSA-Talk.pdf}, 2022.
\newblock Accessed: 2022-03-17.

\bibitem{hartig2009executing}
Olaf Hartig, Christian Bizer, and Johann-Christoph Freytag.
\newblock Executing sparql queries over the web of linked data.
\newblock In {\em The Semantic Web-ISWC 2009: 8th International Semantic Web Conference, ISWC 2009, Chantilly, VA, USA, October 25-29, 2009. Proceedings 8}, pages 293--309. Springer, 2009.

\bibitem{hayes1983building}
Frederick Hayes-Roth, Donald~A Waterman, and Douglas~B Lenat.
\newblock {\em Building expert systems}.
\newblock Addison-Wesley Longman Publishing Co., Inc., 1983.

\bibitem{hendler2009tonight}
James~A Hendler.
\newblock Tonight’s dessert: Semantic web layer cakes.
\newblock In {\em European Semantic Web Conference}, pages 1--1. Springer, 2009.

\bibitem{hogan2020semantic}
Aidan Hogan.
\newblock The semantic web: Two decades on.
\newblock {\em Semantic Web}, 11(1):169--185, 2020.

\bibitem{iglesias1998analysis}
Carlos~A Iglesias, Mercedes Garijo, Jos{\'e}~C Gonz{\'a}lez, and Juan~R Velasco.
\newblock Analysis and design of multiagent systems using mas-commonkads.
\newblock In {\em Intelligent Agents IV Agent Theories, Architectures, and Languages: 4th International Workshop, ATAL'97 Providence, Rhode Island, USA, July 24--26, 1997 Proceedings 4}, pages 313--327. Springer, 1998.

\bibitem{iglesias2023comparison}
Ana Iglesias-Molina, Kian Ahrabian, Filip Ilievski, Jay Pujara, and Oscar Corcho.
\newblock Comparison of knowledge graph representations for user consumption scenarios.
\newblock In {\em International Semantic Web Conference (ISWC) Research Track}, 2023.

\bibitem{ilievski2020kgtk}
Filip Ilievski, Daniel Garijo, Hans Chalupsky, Naren~Teja Divvala, Yixiang Yao, Craig Rogers, Rongpeng Li, Jun Liu, Amandeep Singh, Daniel Schwabe, and Pedro Szekely.
\newblock Kgtk: a toolkit for large knowledge graph manipulation and analysis.
\newblock In {\em International Semantic Web Conference}, pages 278--293. Springer, 2020.

\bibitem{ilievski2021cskg}
Filip Ilievski, Pedro Szekely, and Bin Zhang.
\newblock Cskg: The commonsense knowledge graph.
\newblock In {\em Extended Semantic Web Conference (ESWC)}, 2021.

\bibitem{jain2010ontology}
Prateek Jain, Pascal Hitzler, Amit~P Sheth, Kunal Verma, and Peter~Z Yeh.
\newblock Ontology alignment for linked open data.
\newblock In {\em International semantic web conference}, pages 402--417. Springer, 2010.

\bibitem{kautz2022third}
Henry Kautz.
\newblock The third ai summer: Aaai robert s. engelmore memorial lecture.
\newblock {\em AI Magazine}, 43(1):105--125, 2022.

\bibitem{kazman2000atam}
Rick Kazman, Mark Klein, and Paul Clements.
\newblock {\em ATAM: Method for architecture evaluation}.
\newblock Carnegie Mellon University, Software Engineering Institute Pittsburgh, PA, 2000.

\bibitem{kendall2019ontology}
Elisa~F Kendall and Deborah~L McGuinness.
\newblock {\em Ontology engineering}.
\newblock Morgan \& Claypool Publishers, 2019.

\bibitem{khatri2010designing}
Vijay Khatri and Carol~V Brown.
\newblock Designing data governance.
\newblock {\em Communications of the ACM}, 53(1):148--152, 2010.

\bibitem{lan2022semantic}
Gongjin Lan, Ting Liu, Xu~Wang, Xueli Pan, and Zhisheng Huang.
\newblock A semantic web technology index.
\newblock {\em Scientific reports}, 12(1):3672, 2022.

\bibitem{lenat2023getting}
Doug Lenat and Gary Marcus.
\newblock Getting from generative ai to trustworthy ai: What llms might learn from cyc.
\newblock {\em arXiv preprint arXiv:2308.04445}, 2023.

\bibitem{lerer2019pytorch}
Adam Lerer, Ledell Wu, Jiajun Shen, Timothee Lacroix, Luca Wehrstedt, Abhijit Bose, and Alex Peysakhovich.
\newblock Pytorch-biggraph: A large scale graph embedding system.
\newblock {\em Proceedings of Machine Learning and Systems}, 1:120--131, 2019.

\bibitem{li2023trustworthy}
Bo~Li, Peng Qi, Bo~Liu, Shuai Di, Jingen Liu, Jiquan Pei, Jinfeng Yi, and Bowen Zhou.
\newblock Trustworthy ai: From principles to practices.
\newblock {\em ACM Computing Surveys}, 55(9):1--46, 2023.

\bibitem{lobentanzer2022democratising}
Sebastian Lobentanzer, Patrick Aloy, Jan Baumbach, Balazs Bohar, Vincent~J Carey, Pornpimol Charoentong, Katharina Danhauser, Tunca Do{\u{g}}an, Johann Dreo, Ian Dunham, et~al.
\newblock Democratizing knowledge representation with biocypher.
\newblock {\em Nature Biotechnology}, pages 1--4, 2023.

\bibitem{lobentanzer2023democratizing}
Sebastian Lobentanzer, Patrick Aloy, Jan Baumbach, Balazs Bohar, Vincent~J Carey, Pornpimol Charoentong, Katharina Danhauser, Tunca Do{\u{g}}an, Johann Dreo, Ian Dunham, et~al.
\newblock Democratizing knowledge representation with biocypher.
\newblock {\em Nature Biotechnology}, 41(8):1056--1059, 2023.

\bibitem{nakagawa2011aspect}
Elisa~Y Nakagawa, Fabiano~C Ferrari, Mariela~MF Sasaki, and Jos{\'e}~C Maldonado.
\newblock An aspect-oriented reference architecture for software engineering environments.
\newblock {\em Journal of Systems and Software}, 84(10):1670--1684, 2011.

\bibitem{newell1958elements}
Allen Newell, John~Calman Shaw, and Herbert~A Simon.
\newblock Elements of a theory of human problem solving.
\newblock {\em Psychological review}, 65(3):151, 1958.

\bibitem{noy2001ontology}
Natalya~F Noy, Deborah~L McGuinness, et~al.
\newblock Ontology development 101: A guide to creating your first ontology, 2001.

\bibitem{noy2019industry}
Natasha Noy, Yuqing Gao, Anshu Jain, Anant Narayanan, Alan Patterson, and Jamie Taylor.
\newblock Industry-scale knowledge graphs: Lessons and challenges: Five diverse technology companies show how it’s done.
\newblock {\em Queue}, 17(2):48--75, 2019.

\bibitem{gallofre2020data}
Marc~Gallofr{\'e} Oca{\~n}a, Tareq Al-Moslmi, and A.~Opdahl.
\newblock Data privacy in journalistic knowledge platforms.
\newblock In {\em International Conference on Information and Knowledge Management}, 2020.
\newblock URL: \url{https://api.semanticscholar.org/CorpusID:224820106}.

\bibitem{ocana2023software}
Marc~Gallofr{\'e} Oca{\~n}a and Andreas~L Opdahl.
\newblock A software reference architecture for journalistic knowledge platforms.
\newblock {\em Knowledge-Based Systems}, 276:110750, 2023.

\bibitem{otero2015ontology}
Lorena Otero-Cerdeira, Francisco~J Rodr{\'\i}guez-Mart{\'\i}nez, and Alma G{\'o}mez-Rodr{\'\i}guez.
\newblock Ontology matching: A literature review.
\newblock {\em Expert Systems with Applications}, 42(2):949--971, 2015.

\bibitem{paulheim2017knowledge}
Heiko Paulheim.
\newblock Knowledge graph refinement: A survey of approaches and evaluation methods.
\newblock {\em Semantic web}, 8(3):489--508, 2017.

\bibitem{petroni2019language}
Fabio Petroni, Tim Rockt{\"a}schel, Sebastian Riedel, Patrick Lewis, Anton Bakhtin, Yuxiang Wu, and Alexander Miller.
\newblock Language models as knowledge bases?
\newblock In Kentaro Inui, Jing Jiang, Vincent Ng, and Xiaojun Wan, editors, {\em Proceedings of the 2019 Conference on Empirical Methods in Natural Language Processing and the 9th International Joint Conference on Natural Language Processing (EMNLP-IJCNLP)}, pages 2463--2473, Hong Kong, China, November 2019. Association for Computational Linguistics.
\newblock URL: \url{https://aclanthology.org/D19-1250}, \href {https://doi.org/10.18653/v1/D19-1250} {\path{doi:10.18653/v1/D19-1250}}.

\bibitem{piscopo2018models}
Alessandro Piscopo and Elena Simperl.
\newblock Who models the world? collaborative ontology creation and user roles in wikidata.
\newblock {\em Proceedings of the ACM on Human-Computer Interaction}, 2(CSCW):1--18, 2018.

\bibitem{poveda2022lot}
Mar{\'\i}a Poveda-Villal{\'o}n, Alba Fern{\'a}ndez-Izquierdo, Mariano Fern{\'a}ndez-L{\'o}pez, and Ra{\'u}l Garc{\'\i}a-Castro.
\newblock Lot: An industrial oriented ontology engineering framework.
\newblock {\em Engineering Applications of Artificial Intelligence}, 111:104755, 2022.

\bibitem{preece2001evaluating}
Alun Preece.
\newblock Evaluating verification and validation methods in knowledge engineering.
\newblock In {\em Industrial knowledge management: A micro-level approach}, pages 91--104. Springer, 2001.

\bibitem{priem2022openalex}
Jason Priem, Heather Piwowar, and Richard Orr.
\newblock Openalex: A fully-open index of scholarly works, authors, venues, institutions, and concepts.
\newblock {\em arXiv preprint arXiv:2205.01833}, 2022.

\bibitem{qiu2020dictionary}
Qinjun Qiu, Zhong Xie, Liang Wu, and Liufeng Tao.
\newblock Dictionary-based automated information extraction from geological documents using a deep learning algorithm.
\newblock {\em Earth and Space Science}, 7(3):e2019EA000993, 2020.

\bibitem{ramsey1929knowledge}
F.P. Ramsey.
\newblock {Knowledge}.
\newblock In {\em {F.P. Ramsey: Philosophical Papers}}, pages 110--111. Cambridge University Press, 1929.

\bibitem{ravat2019data}
Franck Ravat and Yan Zhao.
\newblock Data lakes: Trends and perspectives.
\newblock In {\em Database and Expert Systems Applications: 30th International Conference, DEXA 2019, Linz, Austria, August 26--29, 2019, Proceedings, Part I 30}, pages 304--313. Springer, 2019.

\bibitem{sabouknowledge}
Marta Sabou, Majlinda Llugiqi, Fajar~J Ekaputra, Laura Waltersdorfer, and Stefani Tsaneva.
\newblock Knowledge engineering in the age of neurosymbolic systems.
\newblock {\em Neurosymbolic AI Journal (under review)}, 2024.

\bibitem{sahlabadi2022lightweight}
Mahdi Sahlabadi, Ravie~Chandren Muniyandi, Zarina Shukur, and Faizan Qamar.
\newblock Lightweight software architecture evaluation for industry: A comprehensive review.
\newblock {\em Sensors}, 22(3):1252, 2022.

\bibitem{salloum2016big}
Salman Salloum, Ruslan Dautov, Xiaojun Chen, Patrick~Xiaogang Peng, and Joshua~Zhexue Huang.
\newblock Big data analytics on apache spark.
\newblock {\em International Journal of Data Science and Analytics}, 1:145--164, 2016.

\bibitem{schade2023how}
Michael Schade.
\newblock {How ChatGPT and Our Language Models Are Developed}.
\newblock \url{https://help.openai.com/en/articles/7842364-how-chatgpt-and-our-language-models-are-developed}, 2023.
\newblock Accessed: 2024-01-05.

\bibitem{schneider2022decade}
Phillip Schneider, Tim Schopf, Juraj Vladika, Mikhail Galkin, Elena Paslaru~Bontas Simperl, and Florian Matthes.
\newblock A decade of knowledge graphs in natural language processing: A survey.
\newblock In {\em AACL}, 2022.
\newblock URL: \url{https://api.semanticscholar.org/CorpusID:252683270}.

\bibitem{schreiber2000knowledge}
August~Th Schreiber, Guus Schreiber, Hans Akkermans, Anjo Anjewierden, Nigel Shadbolt, Robert de~Hoog, Walter Van~de Velde, and Bob Wielinga.
\newblock {\em Knowledge engineering and management: the CommonKADS methodology}.
\newblock MIT press, 2000.

\bibitem{shenoy2021study}
Kartik Shenoy, Filip Ilievski, Daniel Garijo, Daniel Schwabe, and Pedro Szekely.
\newblock A study of the quality of wikidata.
\newblock {\em Journal of Web Semantics}, 2021.

\bibitem{simsek2022knowledge}
Umutcan Simsek, Elias K{\"a}rle, Kevin Angele, Elwin Huaman, Juliette Opdenplatz, Dennis Sommer, J{\"u}rgen Umbrich, and Dieter Fensel.
\newblock A knowledge graph perspective on knowledge engineering.
\newblock {\em SN Computer Science}, 4(1):16, 2022.

\bibitem{song2017building}
Dezhao Song, Frank Schilder, Shai Hertz, Giuseppe Saltini, Charese Smiley, Phani Nivarthi, Oren Hazai, Dudi Landau, Mike Zaharkin, Tom Zielund, et~al.
\newblock Building and querying an enterprise knowledge graph.
\newblock {\em IEEE Transactions on Services Computing}, 12(3):356--369, 2017.

\bibitem{staron2021autosar}
Miroslaw Staron and Miroslaw Staron.
\newblock Autosar (automotive open system architecture).
\newblock {\em Automotive Software Architectures: An Introduction}, pages 97--136, 2021.

\bibitem{steidl2023pipeline}
Monika Steidl, Michael Felderer, and Rudolf Ramler.
\newblock The pipeline for the continuous development of artificial intelligence models—current state of research and practice.
\newblock {\em Journal of Systems and Software}, 199:111615, 2023.

\bibitem{suarez2011neon}
Mari~Carmen Su{\'a}rez-Figueroa, Asunci{\'o}n G{\'o}mez-P{\'e}rez, and Mariano Fern{\'a}ndez-L{\'o}pez.
\newblock The neon methodology for ontology engineering.
\newblock In {\em Ontology engineering in a networked world}, pages 9--34. Springer, 2011.

\bibitem{tamavsauskaite2022defining}
Gyt{\.e} Tama{\v{s}}auskait{\.e} and Paul Groth.
\newblock Defining a knowledge graph development process through a systematic review.
\newblock {\em ACM Transactions on Software Engineering and Methodology}, 2022.

\bibitem{taylor2010software}
Richard~N Taylor, Nenad Medvidović, and Eric~M Dashofy.
\newblock {\em Software architecture: foundations, theory, and practice}.
\newblock John Wiley \& Sons, Inc., 2010.

\bibitem{wqdssearchteam2022wqds}
WDQS~Search Team.
\newblock {WDQS Backend Alternatives: The Process, Details and Results}.
\newblock \url{https://www.wikidata.org/wiki/File:WDQS_Backend_Alternatives_working_paper.pdf}, 2022.
\newblock Accessed: 2022-08-15.

\bibitem{tharani2021much}
Karim Tharani.
\newblock Much more than a mere technology: A systematic review of wikidata in libraries.
\newblock {\em The Journal of Academic Librarianship}, 47(2):102326, 2021.

\bibitem{thornton2019using}
Katherine Thornton, Harold Solbrig, Gregory~S Stupp, Jose~Emilio Labra~Gayo, Daniel Mietchen, Eric Prud’Hommeaux, and Andra Waagmeester.
\newblock Using shape expressions (shex) to share rdf data models and to guide curation with rigorous validation.
\newblock In {\em The Semantic Web: 16th International Conference, ESWC 2019, Portoro{\v{z}}, Slovenia, June 2--6, 2019, Proceedings 16}, pages 606--620. Springer, 2019.

\bibitem{tiddi2023knowledge}
Ilaria Tiddi, Victor De~Boer, Stefan Schlobach, and Andr{\'e} Meyer-Vitali.
\newblock Knowledge engineering for hybrid intelligence.
\newblock In {\em Proceedings of the 12th Knowledge Capture Conference 2023}, pages 75--82, 2023.

\bibitem{tommasini}
Riccardo Tommasini, Filip Ilievski, and Thilini Wijesiriwardene.
\newblock The internet meme knowledge graph.
\newblock In {\em ESWC}, 2023.

\bibitem{van2021modular}
Michael van Bekkum, Maaike de~Boer, Frank van Harmelen, Andr{\'e} Meyer-Vitali, and Annette~ten Teije.
\newblock Modular design patterns for hybrid learning and reasoning systems: a taxonomy, patterns and use cases.
\newblock {\em Applied Intelligence}, 51(9):6528--6546, 2021.

\bibitem{van2019boxology}
Frank Van~Harmelen and Annette Ten~Teije.
\newblock A boxology of design patterns for hybrid learning and reasoning systems.
\newblock {\em Journal of Web Engineering}, 18(1-3):97--123, 2019.

\bibitem{waltersdorfer2023semantic}
Laura Waltersdorfer, Anna Breit, Fajar~J Ekaputra, Marta Sabou, Andreas Ekelhart, Andreea Iana, Heiko Paulheim, Jan Portisch, Artem Revenko, Annette ten Teije, et~al.
\newblock Semantic web machine learning systems: An analysis of system patterns.
\newblock In {\em Compendium of Neurosymbolic Artificial Intelligence}, pages 77--99. IOS Press, 2023.

\bibitem{wielinga1992kads}
Bob~J Wielinga, A~Th Schreiber, and Jost~A Breuker.
\newblock Kads: A modelling approach to knowledge engineering.
\newblock {\em Knowledge acquisition}, 4(1):5--53, 1992.

\bibitem{witschel2020visualization}
Hans~Friedrich Witschel, Charuta Pande, Andreas Martin, Emanuele Laurenzi, and Knut Hinkelmann.
\newblock Visualization of patterns for hybrid learning and reasoning with human involvement.
\newblock In {\em New Trends in Business Information Systems and Technology: Digital Innovation and Digital Business Transformation}, pages 193--204. Springer, 2020.

\bibitem{yang2006applying}
Dong Yang, Lixin Tong, Yan Ye, and Hongwei Wu.
\newblock Applying commonkads and semantic web technologies to ontology-based e-government knowledge systems.
\newblock In {\em The Semantic Web--ASWC 2006: First Asian Semantic Web Conference, Beijing, China, September 3-7, 2006. Proceedings 1}, pages 336--342. Springer, 2006.

\bibitem{yao2019extensible}
Yixiang Yao, Pedro Szekely, and Jay Pujara.
\newblock Extensible and scalable entity resolution for financial datasets using rltk.
\newblock In {\em Proceedings of the 5th Workshop on Data Science for Macro-modeling with Financial and Economic Datasets}, pages 1--1, 2019.

\bibitem{zalmout2021all}
Nasser Zalmout, Chenwei Zhang, Xian Li, Yan Liang, and Xin~Luna Dong.
\newblock All you need to know to build a product knowledge graph.
\newblock In {\em Proceedings of the 27th ACM SIGKDD Conference on Knowledge Discovery \& Data Mining}, pages 4090--4091, 2021.

\end{thebibliography}


\end{document}